\documentclass[ida]{iosart2x}

\pubyear{0000}
\volume{0}
\firstpage{1}
\lastpage{1}
\usepackage{float}
\usepackage{hyperref}
\hypersetup{
    colorlinks=true,
    linkcolor=blue,
    filecolor=magenta,      
    urlcolor=cyan,
    }
\usepackage[group-separator={,}]{siunitx}

\begin{document}

\makeatletter
\let\put@numberlines@box\relax
\makeatother

\begin{frontmatter}

\title{Identifying Relevant Features of CSE-CIC-IDS2018 Dataset for the Development of an Intrusion Detection System}
\runningtitle{Identifying Relevant Features of CSE-CIC-IDS2018 Dataset for the Development of an Intrusion Detection System}


\author[A]{\inits{N.}\fnms{László} \snm{Göcs}\ead[label=e1]{gocs.laszlo@nje.hu}%
\thanks{Corresponding author. \printead{e1}.}},
\author[B]{\inits{N.N.}\fnms{Zsolt Csaba} \snm{Johanyák}\ead[label=e2]{johanyak.csaba@nje.hu}}
\address[A]{Department of Information Technology, \orgname{John von Neumann University, GAMF Faculty of Engineering and Computer Science},
Izsáki út 10, \cny{Kecskemét}\printead[presep={\\}]{e1}}
\address[B]{Department of Information Technology, \orgname{John von Neumann University, GAMF Faculty of Engineering and Computer Science},
Izsáki út 10, \cny{Kecskemét}\printead[presep={\\}]{e2}}

\begin{abstract}
Intrusion detection systems (IDSs) are essential elements of IT systems. Their key component is a classification module that continuously evaluates some features of the network traffic and identifies possible threats. Its efficiency is greatly affected by the right selection of the features to be monitored. Therefore, the identification of a minimal set of features that are necessary to safely distinguish malicious traffic from benign traffic is indispensable in the course of the development of an IDS. This paper presents the preprocessing and feature selection workflow as well as its results in the case of the CSE-CIC-IDS2018 on AWS dataset, focusing on five attack types. To identify the relevant features, six feature selection methods were applied, and the final ranking of the features was elaborated based on their average score. Next, several subsets of the features were formed based on different ranking threshold values, and each subset was tried with five classification algorithms to determine the optimal feature set for each attack type. During the evaluation, four widely used metrics were taken into consideration.
\end{abstract}

\begin{keyword}
\kwd{dataset preprocessing}
\kwd{dimension reduction}
\kwd{feature selection}
\kwd{classification}
\kwd{Python}
\kwd{CE-CIC-IDS2018}
\end{keyword}

\end{frontmatter}


\section*{1 Introduction}
Nowadays, only automated tools called Intrusion Detection Systems (IDSs) are capable of efficiently detecting attacks against IT systems. They continuously monitor and evaluate the parameters of network packages. An IDS is a software or hardware solution that can detect out-of-the-ordinary packages and activities capable of damaging the computer or even the network. An IDS device monitors traffic passing through network interfaces. As soon as it detects malicious activity, it sends an alarm message to a pre-configured monitoring system that can prevent further attacks by re-configuring network devices such as security appliances or traffic controllers. The IDS is often deployed at the boundary of the trusted network, sometimes even outside the firewall.

Intrusion detection systems can be categorized in several ways, e.g., by the intrusion detection approach used (anomaly-based or signature-based), by the type of system protected (host, network, hybrid), by the IDS architecture (centralized, distributed), by the source of data used for analysis (network packages, system analysis), by the level of service provided after attack detection (active, passive), and by the timing of analysis (continuous, time interval) \cite{lgocs_team2015}.

The results being reported in this paper were obtained in the course of an investigation focusing on anomaly-based intrusion detection systems (also called Behavior-based IDSs – BIDSs). These systems operate in two modes (learning and detection). In the learning mode, the system is fed with sensor data that contain typical (normal) network and malicious (attack) data. The classification unit is trained and tested based on the labels associated with the data records.  In detection mode, the fully trained classification module aims to determine whether the current activity is harmful to the system or not. The anomaly-based approach has the advantage of being able to adapt quickly and dynamically to unknown attack types. BIDSs can be classified into three main categories based on the way they process data, namely statistical-based, knowledge-based, and computational intelligence-based \cite{lgocs_team2016}. 

The classification module is the most important component of an IDS. Its efficiency and speed are affected to a great extent  by the right selection of the features being monitored and used for the classification.  The main focus of the research reported in this paper was on determining these features in the case of several network datasets containing normal data as well as data related to different attack types. For each attack type, the rank of each feature was determined based on the average score obtained from the results of the application of several feature selection methods. Next, different classification methods were used to evaluate a series of rank threshold values to determine the optimal threshold and feature set for each attack type. 

The rest of the paper is organized as follows. Section 2 contains a literature review. Section 3 presents the used datasets and the preprocessing steps. The applied feature selection methods are described in Section 4. Section 5 describes the classification algorithms used in the course of the evaluation while the results are discussed in Section 6. The conclusions are drawn in Section 7.

\section*{2 Related works}

The sample data based development possibilities of IDS classification modules have been intensively investigated in the last decade. Kurniabudi et al. \cite{Stiawan_ieee} used Information Gain (IG) method to rank and cluster features of the CICIDS-2017 dataset and then applied Random Forest (RF), Bayes Net (BN), Random Tree (RT), Naive Bayes (NB) and J48 classification algorithms to select the features, which yielded good classification results.

Rahman et al.  \cite{Rahman_Springer}  performed AWID dataset analysis using Support Vector Machine (SVM) and C4.5  as feature selection methods using artificial neural networks (ANN) based classification achieving 99.95\% accuracy.
Javadpour et al. \cite{Javadpour_ieee} used Pearson Linear Correlation and IG to select the features of the KDD99 dataset and CART,  ANN, Decision Tree, and Random Forest (RF) algorithms for classification. They obtained the best results  (99.98\% accuracy) using the neural network method.
Taher et al. (2019) used correlation and Chi-square-based techniques as feature selection methods for the NSL-KDD dataset, followed by ANN and SVM classification algorithms achieving 94.02\% recognition rate \cite{Taher_ieee}.

Kocher et al. used the Chi-square approach for dimensionality reduction on the UNSW-NB15 dataset followed by k-Nearest Neighbors (KNN), Stochastic Gradient Descent (SGD), Random Forest, Logistic Regression (LR) and Naive Bayes (NB) algorithms for classification, and resulting in a  classifier accuracy of  99.64\%  \cite{Kocher_ssrn}.
Alkasassbeh \cite{Alkasassbeh_arxiv} used BayesNet, MLP, and SVM machine learning methods, and IG, ReliefF (RF), and Genetic Search (GS) for feature selection. The best accuracy (99.9\%) was achieved with BayesNet and GS.

Thaseen et al. used the Chi-square approach to select features of the NSL KDD dataset and did the classification with an SVM classifier. The proposed model resulted in a high detection rate and a low false alarm rate \cite{Thaseen_elseiver}.
Awotunde et al. compared NSL-KDD and UNSW-NB15 datasets using hybrid rule-based feature selection and the DFFNN deep learning algorithm, with a recognition rate of 99.0\% for NSL-KDD and 98.9\% for UNSW-NB15 \cite{Awotunde_hindawi}.

Sasan et al. applied the J48 and Classification \& regression Trees (CART) methods for the NSL-KDD dataset using 29 features and achieved 88.23\% accuracy \cite{Sasan_ijcs}. However, the article does not describe how the 29 features were selected.
Biswas presented a comparison of feature selection methods (CFS, IGR, PCA) and classifier algorithms (NB, SVM, DT, NN, k-NN) on the NSL-KDD dataset, which shows that the k-NN classifier performs better than the others and among the feature selection methods, IGR feature selection method is better \cite{Biswas_ijpa}.

Shaukat et al.  investigated the CICIDS-2017 dataset using CFS and Naive Bayes feature selection methods with MLP and IBK algorithms, which showed that IBK is more accurate than MLP \cite{Ali_ieee}.
Malhotra et al.   used Naive Bayes, Bayes Net, Logistic, Random Tree, Random Forest, J48, Bagging, OneR, PART, and ZeroR classifiers for the analysis of the NSL-KDD dataset, out of which Random Forest, Bagging, PART, and J48 were the best four in terms of model construction time. However, Random Tree achieved good accuracy in a short time without using feature selection and dimension reduction methods \cite{Malhotra_ijcnis}.

Krishnaveni et al. used IG, Chi-square, Gain Ratio, Symmetric Uncertainty, and Relief methods for feature selection for Real-Time Honeypot, NSL-KDD, and Kyoto datasets, and also used SVM, Naive Bayes, Logistic Regression, and Decision Tree classification algorithms \cite{Krishnaveni_springer}.
Kumar et al. used CFS, IGF, and GR methods for the feature selection in the case of the NSL-KDD dataset and applied Naive Bayes, J48, and RepTree algorithms for classification. The feature subset identified by GR and Ranker improved the proposed Naive Bayes classification \cite{Kumar_ijca}.

Pattawaro and Polprasert utilized a feature selection method based on attribute ratio (AR) for the NSL-KDD dataset combined with k-Means clustering and XGBoost classification. The proposed model achieved an accuracy of 84.41\% \cite{Pattawaro_ieee}.
Tohari et al. worked with k-Nearest Neighbor (k-NN), SVM, and Naive Bayes classifiers on the KDD Cup99, Kyoto 2006, and UNSW-NB15 datasets [18], where the best performance is achieved by SVM with 99.9291\% accuracy and 0\% false positive rate \cite{Ahmad_icic}.

\section*{3 Preprocessing the datasets}
The classification module is the core module of an IDS. Usually, it is developed using one or more sample datasets and applying statistical or machine learning techniques. These datasets contain an immense amount of data describing normal (benign) and malicious traffic. The raw data obtained from the sensors usually have to undergo several preprocessing steps until it can be used for the training of the classification module. These steps can be divided into three main phases: data cleaning, data transformation, and data reduction. In the following subsections, after a short introduction of the available and used datasets the preprocessing applied in course of this investigation is presented in detail. 

\subsection*{3.1 Datasets}
The first sample dataset used for IDS training purposes was the famous KDD’99 \cite{Aggarwal_Elsevier}, which has served later as a starting point for the development of several IDS solutions. It contains information about simulated traffic corresponding to normal activities and several attack types (DOS, guesspassword, buffer overflow, remote FTP, synflood, Nmap, rootkit). Subsequently, some other datasets have been also created containing samples of new attack types as well as some additional features. The most relevant datasets are presented in Table~\ref{table:IDS_Datasets}. The second column presents the attack types covered by the given dataset.

\begin{table*}[!ht]
\renewcommand{\arraystretch}{2}
\caption{IDS Datasets}
\label{table:IDS_Datasets}
\begin{tabular}{p{0.25\textwidth}p{0.65\textwidth}}
\hline
\multicolumn{1}{c}{\textbf{Dataset}} & \multicolumn{1}{c}{\textbf{Attack types included}} \\ 
\hline

CSE-CIC-IDS2018 \newline on AWSnet \cite{Basnet_jisis}     & Bruteforce attack, DoS attack, Web attack,   Infiltration attack, Botnet attack, DDoS+PortScan \\
CIC-IDS2017 \newline \cite{Sharafaldin_icissp} \cite{Chen_jcsm} \cite{Sharafaldin_sn} & botnet (Ares), cross-site-scripting, DoS (executed,   through Hulk, GoldenEye, Slowloris, and Slowhttptest), DDoS (executed through   LOIC), heartbleed, infiltration, SSH brute force, SQL injection     \\
CIDDS-001 \cite{Ring_jiw} \cite{Ring_acpi} & DoS, port scans (ping-scan, SYN-Scan), SSH brute 
force \\
CIDDS-002 \cite{Ring_jiw} \cite{Ring_acpi} & port scans (ACK-Scan, FIN-Scan, ping-Scan, UDP-Scan,   SYN-Scan) \\
UNSW-NB15 \cite{Moustafa_milcis} \cite{Moustafa_isj} & backdoors, DoS, exploits, fuzzers, generic, port   scans, reconnaissance, shellcode, spam, worms  \\
UGR’16 \cite{Macia_cs} & botnet (Neris), DoS, port scans, SSH brute force,   
spam \\
TUIDS \cite{Bhuyan_ijns} & botnet (IRC), DDoS (Fraggle flood, Ping flood, RST   flood, smurf ICMP flood, SYN flood, UDP flood), port scans (e.g. FIN-Scan,   NULL-Scan, UDP-Scan, XMAS-Scan), coordinated port scan, SSH brute force \\ \hline
\end{tabular}
\end{table*}

The dataset used in course of the research reported in this paper is the CSE-CIC-IDS2018 on AWS \cite{Basnet_jisis} that was created by the Canadian Institute for Cybersecurity lab. This dataset was chosen because it is one of the most recent ones and meets all the criteria (e.g. total traffic, many attacks, labeling) required for the research. The dataset includes seven different attack types, i.e., Intrusion, Brute Force, Heartbleed, Botnet, DoS, DDoS, web attacks, and network infiltration. The infrastructure used for the simulation of the attacks consisted of 50 machines while the victim organization consisted of 5 departments, 420 machines, and 30 servers. The dataset contains a record of the network traffic and system logs of each machine and 80 attributes extracted from the recorded traffic using CICFlowMeter-V3 \cite{Lashkari_icissp}. 

The dataset actually consists of several files. The files selected for investigation  and the attack types covered by them are presented in Table \ref{table:selected_datasets}. The preprocessing started with merging these files. The three stages of the preprocessing workflow are presented in the next three subsections.

\begin{table*}[!ht]
\renewcommand{\arraystretch}{2}
\caption{Selected datasets}
\label{table:selected_datasets}
\begin{tabular}{p{0.5\textwidth}p{0.15\textwidth}}
\hline
\textbf{File name} & \textbf{Attack types}  \\ \hline
Wednesday-14-02-2018 TrafficForML CICFlowMeter.csv & FTP-BruteForce \newline 
                                                     SSH-BruteForce \\
Thursday-22-02-2018 TrafficForML CICFlowMeter.csv  & Brute Force –Web \newline 
                                                     Brute Force –XSS \newline 
                                                     SQL Injection\\
Friday-23-02-2018 TrafficForML CICFlowMeter.csv    & Brute Force –Web \newline 
                                                     Brute Force –XSS \newline 
                                                     SQL Injection \\ \hline
\end{tabular}
\end{table*}

\subsection*{3.2 Data cleaning}

\begin{figure*}[!ht]
    \centering
    \includegraphics[width=1\textwidth]{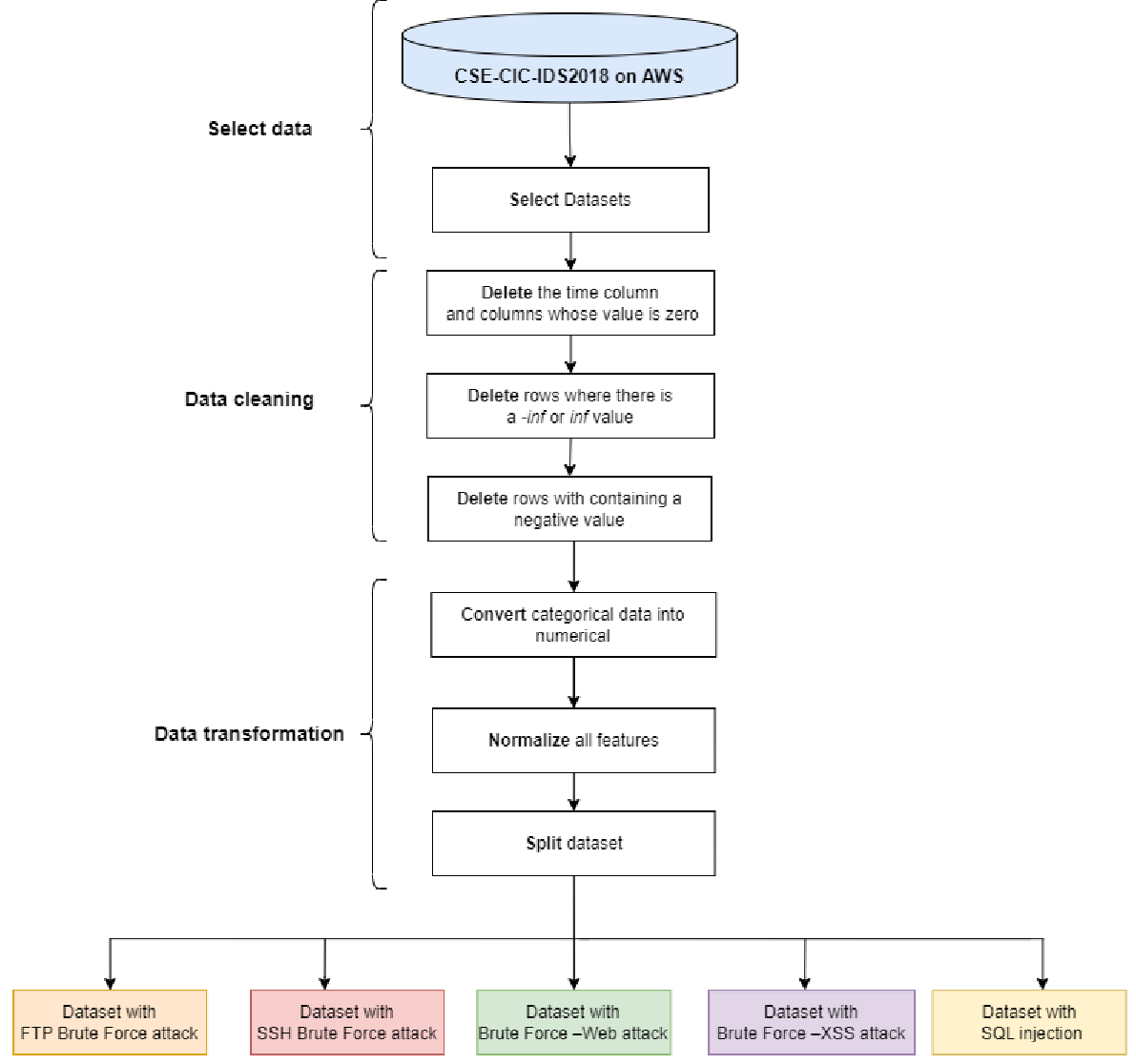}
    \caption{Dataset preprocessing workflow}  
    \label{fig:preprocessing_workflow}
\end{figure*}

In order to create a proper dataset to train and test the model one has to preprocess the raw data. The preprocessing workflow starts with data cleaning, which usually includes deleting rows (records) containing invalid or missing data, deleting one-valued columns (e.g. columns where all values are zero), as well as deleting features (columns) that are a-priory known to be irrelevant regarding the classification. The main steps are illustrated in Fig.~\ref{fig:preprocessing_workflow}. This could already achieve some dimensionality reduction, which can provide various advantages that will be discussed later. 

For the current analysis, the time parameter is not needed, nor are the columns where all values are zero since they do not influence the output, i.e. the value of the last column. The data cleaning step resulted in 69 remaining columns after deleting 11 columns out of the original 80 ones. The next step was to delete rows containing invalid values. Therefore, first, the rows with the values $inf$ and $-inf$ were deleted, and then rows with negative values in the dataset were also deleted. With these operations, the cleaning of the dataset was completed.

\subsection*{3.3 Data transformation}

In course of this research, the data transformation phase comprised three operations, i.e. the transformation of categorical data into numerical, normalization, and splitting of the dataset. To perform further calculations, all non-numerical elements of the dataset were converted into numbers. It was carried out in the case of each categorical column by assigning a number to each category. For example, each occurrence of the string "FTP-BruteForce" was replaced by the value 1. The order of the values was determined arbitrarily not taking into consideration any conceptual distance metrics mainly because later on the dataset was split into subsets containing only data corresponding to one attack type and normal traffic.

Data normalization is a common practice in machine learning, which consists of converting numeric columns to a common scale. In machine learning, some feature values are several times greater than others. Thus the features with higher values could dominate the learning process. However, it does not mean that these variables are more important in predicting the model output. Data normalization converts multiple scaled data to the same scale. After normalization, all variables have similar scale-related effects on the model, which improves the stability and performance of the learning algorithm. There are several types of normalization techniques. The simplest and most used type is the min-max scaling (\ref{equation:minmax_normalization}), which rescales a feature into the fixed range [0,1] by subtracting the minimum value of the feature ($x_{min}$) from the current value ($x$) and then dividing the result by the range.

\begin{equation}
\label{equation:minmax_normalization}
X_{new}=\frac{x-x_{min}}{x_{max}-x_{min}}
\end{equation}

As a final step in the preprocessing, the selected datasets were split so that each resulting file contained only records corresponding to one attack type as well as records describing normal traffic. This operation resulted in five data files (see Fig.~\ref{fig:preprocessing_workflow}). Some of their features are presented in Table~\ref{table:Datasets_preproc}.

\begin{table}[!ht]
\renewcommand{\arraystretch}{2}
\caption{Datasets generated during preprocessing}
\label{table:Datasets_preproc}

\begin{tabular}{p{0.28\columnwidth}p{0.25\columnwidth}p{0.31\columnwidth}}
\hline
\textbf{File name} &  \textbf{Number of rows} & \textbf{Number of columns}\\ 
\hline
dataset-ftp.csv     & \hfil \num{857162}   & \hfil 69 \\
dataset-ssh.csv     & \hfil \num{851397}   & \hfil 69 \\
dataset-web.csv     & \hfil \num{2085515} & \hfil 69 \\ 
dataset-xss.csv     & \hfil \num{2085134} & \hfil 69 \\ 
dataset-sql.csv     & \hfil \num{2084991} & \hfil 69 \\ 
\hline
\end{tabular}
\end{table}

\subsection*{3.4 Data reduction}
The data reduction phase focuses on feature selection and dimensionality reduction, which can provide several advantages. One of the most important benefits is that many data mining algorithms work better when the number of dimensions - the number of attributes (columns) in the data - is smaller. This is partly because dimensionality reduction eliminates irrelevant attributes and reduces noise. Another advantage is that it can lead to a more comprehensible model because there will be fewer features in it. In addition, the reduced amount of data requires less storage space and less time for its processing. 

There are many software tools that can be used to preprocess and analyze large datasets (e.g. Matlab, SPSS, Orange, Python). In the course of this research, we opted for the usage of the Python programming language because it is free of charge and its modules and libraries can produce fast and efficient results. The applied methods and the obtained results are presented in the next section.

\section*{4 Feature selection}

Feature selection focuses on finding the most relevant attributes, which can be used to carry out an effective classification or prediction \cite{muhi_mtk}\cite{viharos_elseiver}\cite{dobjan_ieee}. It contributes to the reduction of the dimensionality of the problem and so to the decrease of the resource requirements (storage, computation) as well as it can improve the performance of machine learning algorithms \cite{Chauhan_kdn}, i.e. faster training reduced over-fitting, and sometimes better prediction power.  Although this approach may seem to lead to loss of information, this is not the case when redundant or irrelevant information is present. Redundant features are copies of most or all of the information found in one or more other attributes or they can be obtained as a combination of other features. 

Irrelevant attributes contain almost no information that is useful for the data mining task to be performed. Redundant and irrelevant features can reduce the accuracy of the classification and the quality of the clusters discovered. While some irrelevant and redundant attributes can be removed immediately by common sense or professional knowledge, the selection of the best subset of at-tributes often requires a systematic approach. 

The ideal approach to feature selection is trying all possible subsets of features as input to the data mining algorithm used and then selecting the subset that produced the best results. However, this technique would require an enormous amount of time and computational power. Therefore, several other methods have been developed for this purpose primarily based on statistical assumptions. There are three basic approaches to feature selection.

\begin{itemize}
\item \emph{Wrapper methods}: the feature selection algorithm uses the learning method as a subroutine with the computational burden of invoking the learning algorithm to evaluate each subset of features. It finds the best feature set for a given type of machine learning algorithm.
\item \emph{Embedded methods}: the machine learning algorithm decides what attributes to use and what features to ignore.
\item \emph{Filter methods}: the features are selected before the data mining algorithm is run, using a method that is independent of the data mining task.
\end{itemize}

All the feature selection methods presented in the next subsections and used in course of this investigation belong to the group of filter methods. Their advantage is that their time complexity is the lowest among the three groups, and usually, after their application, the machine learning algorithm is less prone to over-fitting.

\subsection*{4.1 Information Gain}

\begin{eqnarray}
H(Y)=-\sum p(y)log_{2}(p(y)),
\end{eqnarray}

where $p(y)$ is the marginal probability density function for the random variable $Y$. If the observed values of $Y$ in the training dataset are partitioned according to the values of a second feature $X$, and the entropy of $Y$ with respect to the partitions induced by $X$ is less than the entropy of $Y$ prior to partitioning, then there is a relationship between features $Y$ and $X$. Thus the entropy of $Y$ after observing $X$ is

\begin{eqnarray}
H(Y|X)=\sum p(x)\sum p(y|x)log_{2}(p(y|x)),
\end{eqnarray}

where $p(y|x)$ is the conditional probability of $y$ given $x$. Given the entropy as a criterion of impurity in a dataset, one can define a measure reflecting additional information about $Y$ provided by $X$ that represents the amount by which the entropy of $Y$ decreases. This measure, an indicator of the dependency between $X$ and $Y$, is known as information gain (\ref{equation:information_gain}).

\begin{eqnarray}
\label{equation:information_gain}
IG(X,Y)=H(Y)-H(Y|X)=H(X)-H(X|Y).
\end{eqnarray}

IG is a symmetrical measure. The method provides an orderly classification of all the features, and then a threshold is required to select a certain number of them according to the order obtained. A weakness of the IG criterion is that it is biased in favor of features with more values even when they are not more informative \cite{Karegowda_ijitkm}.

\subsection*{4.2 Gain Ratio}
The information gain method prefers to select attributes having a large number of values, which led to the development of the feature selection method gain ratio (GR) which is a modification of the information gain aiming the decrease its bias. Originally developed for decision trees GR takes the number and size of the branches into account when choosing an attribute. It improves the evaluation given by information gain by taking into account the number of splits in the feature, i.e., how equally they are distributed \cite{Priyadarsini_ictact}. GR reflects the relevance of each feature the higher its value is the higher the influence of the feature is. Gain ratio is calculated by formula (\ref{equation:gain_ratio})

\begin{equation}
\label{equation:gain_ratio}
GR(X,Y)  = \frac{IG(X,Y)}{SplitInfo},
\end{equation}

\begin{equation}
SplitInfo = - \sum_{}^{}p(i) * log_2(p(i)),
\end{equation}

where $p(i)$ represents the proportion of instances that fall into a particular split of the feature \cite{Pasha_ieee}.

\subsection*{4.3 Relief}
The Relief method calculates a weight value for each feature ($W_j$= weight of feature ‘j’) that can be used to estimate the quality or relevance of the feature \cite{Urbanowicz_elseiver}. The weight vector is initialized with zero values, and it is updated using an iterative approach. The algorithm takes a random sample of $m$ elements from the dataset. In the case of each instance of the sample ($R_i$) it searches for the closest instance that belongs to the same class ($H_i$) and for the closest instance that belongs to the other class ($M_i$). Next, the value of each feature weight is updated by the formula (\ref{equation:relief}).

\begin{equation}
\label{equation:relief}
W_{j}  = W_{j}-\frac{D(R_{ij},H_{ij})}{m}+\frac{D(R_{ij},M_{ij})}{m},
\end{equation}

where $X_{ij}$ is the $j$th feature of the $i$th instance (here $X$ can be $R$, $H$, or $M$). The function $D$ is defined as:

\begin{equation}
D(x,y)=\left\{\begin{matrix}
0 & if\: x=y\\ 
1 & otherwise
\end{matrix}\right.
\end{equation}

The shortcoming of Relief is that it does not identify redundant features, and it can be used only in the case of binary classification problems.

\subsection*{4.4 Symmetric Uncertainty}
Symmetric uncertainty (SU) can be used to calculate the rank of features for feature selection by calculating the relevance between the feature and the class label. A feature with a high SU value gets high importance \cite{Singh_srp}. SU is calculated by normalizing the double value of IG to the sum of the entropies of the two variables.

\begin{eqnarray}
SU(X,Y)=\frac{2xIG(X,Y)}{H(X)+H(Y)},
\end{eqnarray}

where $H(X)$ and $H(Y)$ are the entropy values of variables $X$ and $Y$, while 
$IG(X,Y)$ is the information gain related to the variables $X$ and $Y$, respectively \cite{Bakhshandeh_springer}.

\subsection*{4.5 Chi-Squared}
The Chi-squared test is a statistical hypothesis test that measures divergence from the expected distribution if one assumes that the feature occurrence is actually independent of the class value \cite{forman2003extensive}. The higher the value of chi-squared, the more relevant the feature with respect to the class is. Its calculation is based on the formula (\ref{equation:chi_squared}) \cite{bolon_springer}.

\begin{eqnarray}
\label{equation:chi_squared}
\chi^{2}=\sum_{i=1}^{n_{I}}\sum_{j=1}^{n_c}\frac{[A_{ij}-\frac{R_{i}B_{j}}{N}]^{2}}{\frac{R_{i} B_{j}}{N}},
\end{eqnarray}

where $n_{I}$ is the number of intervals, $n_c$ is the number of classes, $N$ is the total number of instances, $Aij$ the number of instances in the interval $i$ and class $j$, $Ri$ denotes the number of instances in the interval $i$, and $Bj$ the number of instances in class $j$. The Chi-squared test based evaluation was developed for discrete variables. Therefore in the case of continuous features before its application a discretization has to be carried out.

\subsection*{4.6 ANOVA}
Analysis of Variance (ANOVA) is a statistical analysis technique used to compare the means of multiple groups to determine if there is a significant difference between them. Similar to the Chi-squared approach a discretization is necessary before its usage \cite{Kumar_elseiver}. The key idea of ANOVA is to compare the total variance of the data to the variation within the groups and the variation between the groups. 
The within-group sum of squares (SSW) (\ref{equation:SSW}) measures the variation within the groups. It is defined as
\begin{equation}
\label{equation:SSW}
SSW=\sum_{i=1}^{k} [(n_i - 1) * SS(i)],
\end{equation}
where $n_i$ is the number of instances in group $i$, and $SS(i)$ is the variance of group $i$.

The Sum of Squares between groups (SSB) (\ref{equation:SSB}) measures the variation between the means of the groups. It is defined as
\begin{equation}
\label{equation:SSB}
SSB = K * \sum (x-\bar{\bar{x}})^2
\end{equation}

where $K$ is the number of groups, $\bar{x_i}$ is the mean of group $i$, and $\bar{\bar{x}}$ is the mean of all instances.

The total sum of squares (SST) (\ref{equation:SST}) is
\begin{equation}
\label{equation:SST}
SST=SSW+SSB
\end{equation}

The null hypothesis of ANOVA is that all groups have the same mean, i.e., the values of the investigated feature do not have effect on the final class. The alternative hypothesis is that at least one group has a different mean. The null hypothesis is tested by the help of the F-ratio, which is the ratio of SSB to SSW (\ref{equation:Fcalculated}). An F-ratio higher than a threshold value (called critical F-value) indicates that there is a significant difference between the means of the groups, and so the null hypothesis can be rejected.

\begin{equation}
\label{equation:Fcalculated}
F=SSB/SSW
\end{equation}

The critical F-value is that value of the F-distribution, which is defined by the degrees of freedom for the numerator ($df_{SSB}=K-1$ and the degrees of freedom for the denumerator ($df_{SSW}=N-K$). Here $N$ is the number of instances.

\subsection*{4.7 Feature selection results}
\begin{figure*}[!ht]
    \centering
    \includegraphics[width=1\textwidth]{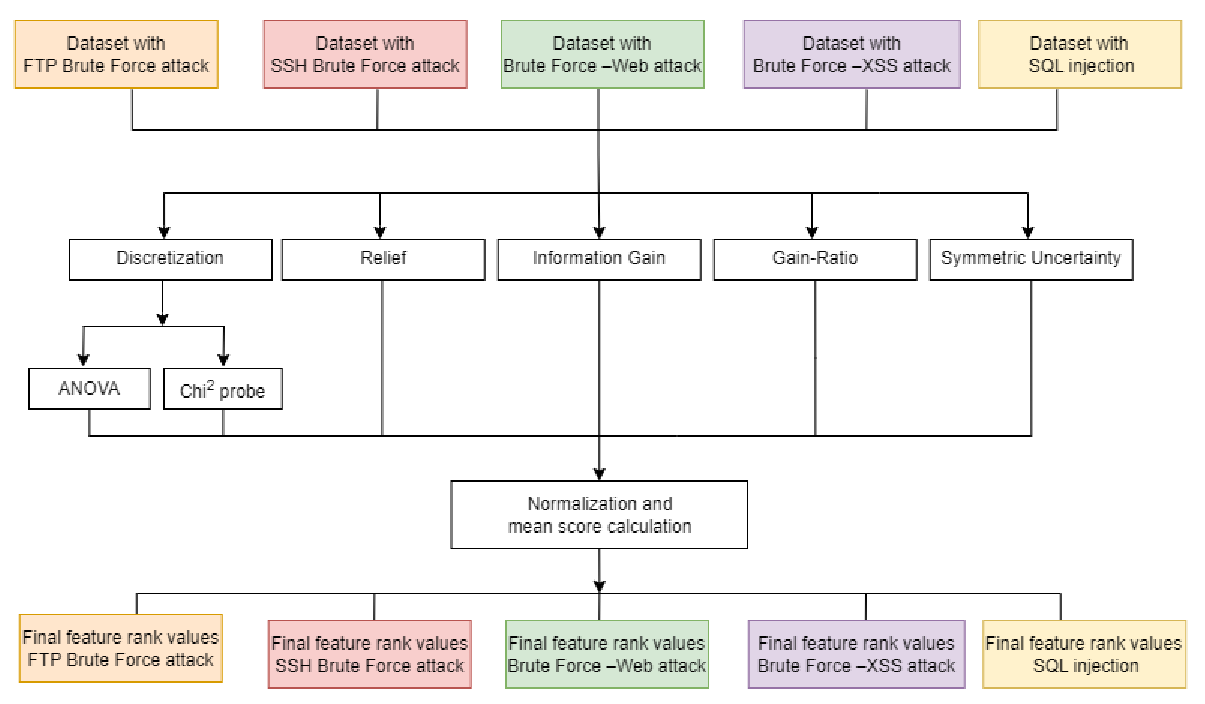}
    \caption{Feature selection workflow}
    \label{fig:feature_select}
\end{figure*}

The six feature selection methods presented in the previous section were applied for all five datasets using 30 university lab computers as well as the ELKH cloud services \cite{Heder_2022}. The feature selection workflow is presented in Fig. \ref{fig:feature_select}. All the necessary program elements were implemented in Python \cite{Bernard_springer} \cite{McKinney_pyt}. Although several tasks were performed in parallel the whole process took more than two months.

In the case of each dataset and each method, the feature score values obtained at the end of the feature selection process were normalized. Next, the final feature score was calculated in the case of each dataset separately as the mean of the normalized scores. The detailed results can be found in the Appendix in Tables A1 - A5.

Finally, five feature ranking threshold values were set starting from 0.35 and increasing by a step of 0.5. In the case of each value, we selected the features whose score was higher than the threshold. The results are presented in Table~\ref{table:feature_ranking}. In the case of each attack type a separate list of relevant features was identified. Each feature is represented by its ordinal value. Each row of the table contains those features whose score was greater or equal to the threshold given in the first cell. Starting from the second column each column represents an attack type. 

Having the selected feature groups we continued our investigation by applying different classification methods that will be presented in the following section.

\renewcommand{\arraystretch}{2}
\begin{table*}[!ht]
\caption{Feature selection based on threshold values}
\label{table:feature_ranking}
\begin{tabular}{p{0.1\textwidth}p{0.14\textwidth}p{0.14\textwidth}p{0.14\textwidth}p{0.14\textwidth}p{0.14\textwidth}}
\hline
\textbf{Ranking threshold} &\textbf{FTP}  &\textbf{SSH}   &\textbf{WEB}  &\textbf{XSS}  &\textbf{SQL}      \\ \hline
0.35   
& 02, 17, 19, 35, 00, 44, 56, 59 
& 00, 02, 17, 19, 57, 56, 59 
& 16, 20, 10, 49, 66, 67, 35, 38, 56, 64, 34, 27, 07, 09, 11, 14, 15, 50, 25, 60, 62, 02, 17, 19, 37, 63, 06, 33, 55, 18, 58, 04, 05, 53, 54, 03, 21, 22, 23, 24, 52, 32, 65, 57 
& 25, 27, 16, 40, 02, 05, 17, 19, 34, 53, 06, 18, 55, 21, 22, 23, 24, 51, 13, 39, 57, 37, 56, 33, 32, 03, 11, 52, 04, 54, 58
& 39, 43, 47, 10, 15, 05, 26, 53, 56, 25, 02, 17, 19, 35, 16, 18, 27, 28, 34, 06, 23, 30, 55, 29, 21, 22, 24, 57, 37, 11, 14\\
 
0.40  
& 44, 56, 59 
& 56, 59 
& 34, 27, 07, 09, 11, 14, 15, 50, 25, 60, 62, 02, 17, 19, 37, 63, 06, 33, 55, 18, 58, 04, 05, 53, 54, 03, 21, 22, 23, 24, 52, 32, 65, 57  
& 02, 05, 17, 19, 34, 53, 06, 18, 55, 21, 22, 23, 24, 51, 13, 39, 57, 37, 56, 33, 32, 03, 11, 52, 04, 54, 58
& 05, 26, 53, 56, 25, 02, 17, 19, 35, 16, 18, 27, 28, 34, 06, 23, 30, 55, 29, 21, 22, 24, 57, 37, 11, 14\\    

0.45 
& 56, 59    
& 56, 59 
& 02, 17, 19, 37, 63, 06, 33, 55, 18, 58, 04, 05, 53, 54, 03, 21, 22, 23, 24, 52, 32, 65, 57 
& 37, 56, 33, 32, 03, 11, 52, 04, 54, 58 
& 06, 23, 30, 55, 29, 21, 22, 24, 57, 37, 11, 14\\

0.50  
& 56, 59   
& 59  
& 03, 21 ,22, 23, 24, 52, 32, 65, 57 
& 03, 11, 52, 04, 54, 58
& 57, 37, 11, 14\\  

0.55 
& 56, 59    
& 59 
& 57
& 58  
& 11,14\\\hline
\end{tabular}
\end{table*}

\section*{5 Classification algorithms}
Classification methods are used to predict the class of an object instance based on a feature vector. Machine learning-based classification algorithms build models that can learn from labeled datasets and use them to predict the class of new, unseen data points. In this investigation, we used five different classification algorithms representing four main classification groups. These groups are linear models, probabilistic models, tree-based models, and kernel-based models.

Linear models are represented by the Logistic Regression method, which models the probability of a binary outcome using a sigmoid function.

Probabilistic models are represented by the Naive Bayes model, which assumes that the features are independent given the class and uses Bayes' theorem to compute the posterior probabilities of each class.

Tree-based models are represented by two methods: the Decision Tree method, a non-parametric model that recursively partitions the feature space into a tree structure, and the Random Forest method, an ensemble model that uses multiple decision trees and aggregates their predictions to improve performance.

Kernel-based models are represented by the Support Vector Machine (SVM) method, which maps the input data into a high-dimensional feature space and finds a hyperplane that maximally separates the classes.

The following subsections provide a brief description of the classification algorithms mentioned above.

\subsection*{5.1 Logistic Regression}
Logistic Regression (LR) is a linear classification method that estimates the probability of a certain instance belonging to a class (e.g. attack). LR belongs to the family of linear methods and is an alternative to discriminant analysis. Its application prerequisites are less strict than those of discriminant analysis \cite{maalouf2011logistic}. The key idea of Logistic Regression is to calculate a linear combination (see eq. (\ref{equation:logisticregrZ})) of the feature values $X=[x_1, x_2, \ldots, x_n]$ for each observation instance, using a coefficient vector $A=[a_0, a_1, \ldots, a_n]$ that is determined during the classifier's training.

\begin{equation}
  \label{equation:logisticregrZ}
  Z(x)=a_0+a_1*x_1+\ldots+a_n*x_n,
\end{equation}

To determine the probability of belonging to the attack class (class 1)  Logistic Regression applies a sigmoid function to $Z$ that maps $Z$ to the $[0, 1]$ interval.

\begin{equation}
  \label{equation:logisticregrSigm}
  h(z)=\frac{1}{1+e^{-z}},
\end{equation}

Finally, the classifier decides the final class of the observation by comparing the resulting probability value $h$ to a threshold value $h_{tr}$, as shown in equation \ref{equation:logisticregrClass}.
 
\begin{equation}
  \label{equation:logisticregrClass}
  C(h)=\left\{\begin{matrix}
         1 & if\: h>h_{tr}\\ 
         0 & otherwise 
       \end{matrix}\right.
\end{equation}

The threshold value $h_{tr}$ is determined during the training phase of the classifier.

\subsection*{5.2 Naive Bayes}

The Naive Bayes classification method is based on Bayes' theorem of conditional probability. It determines the predicted class of an $X=[x_1, x_2, \ldots, x_n]$ observation using the formula

The Naive Bayes classification method is based on Bayes' theorem of conditional probability. It predicts the class of an observation $X=[x_1, x_2, \ldots, x_n]$ using the formula (\ref{equation:NaiveBayes})

\begin{equation}
  \label{equation:NaiveBayes}
  C(x)=\underset{c_j\in C}{\arg\max}\: P(c_j)\prod_{i=1}^{n}P(x_i|c_j)  
\end{equation}

where $c_j$ is the $j$th class, $x_i$ is the value of the $x_i$th feature, $n$ is the number of features, $P(c_j)$ is the prior probability of class $j$, and $P(x_i|c_j)$ is the conditional probability of the value of $x_i$ given class $c_j$. The prior probability $P(c_j)$ is estimated by the relative frequency of class $j$ in the training sample. 

In the case of categorical features $P(x_i|c_j)$ is estimated by the relative frequency of the value $x_i$ among the training sample elements that belong to class $c_j$. In the case of continuous features $P(x_i|c_j)$ is estimated by the value of the probability density function calculated for $x_i$ taking into consideration the training sample elements that belong to class $c_j$

\begin{equation}
  \label{equation:DensityFunction}
  P(x_i|c_j) = \frac{1}{\sigma_i \sqrt{2\pi}} e^{-\frac{(x_i-\mu_i)^2}{2\sigma^2}},
\end{equation}

where $\mu_i$ is the mean value and $\sigma_i$ is the standard deviation of the $i$th feature  among the training sample elements taken into consideration.

\subsection*{5.3 Support Vector Machine }

Support Vector Machine (SVM) \cite{steinwart_support_2008} is a statistically based supervised classification technique that can be used to efficiently handle high-dimensional data. It creates a multi-dimensional hyperplane that separates the two classes in the case of binary classification problems.
Multiclass problems are reduced to multiple binary classification problems.

If no simple linear separation can be carried out it transforms the data by using so-called kernel functions that calculate the hyperplane in a higher dimension.
The nonlinearity of the hyperplane can also be tuned with the help of the regularization and the gamma
parameters. The value of the regularization parameter describes how much one wants to avoid misclassification in the case of training instances. A high value could result in a more complex hyperplane with a small amount of wrongly classified data points if any. 

In the case of high gamma values, only training instances close to the hyperplane will be considered in the course of its definition.

\subsection*{5.4 Decision Tree}

Decision trees offer an easy-to-interpret and visualize tool for classification. They make the decision based on rules inferred from the feature values of the training sample. Each leaf of the tree corresponds to a class label. At each node, only one feature is taken into consideration and no root-to-leaf path  contains twice the same feature. A Classification tree may also provide a confidence measure regarding the quality of the classification.
The tree is built up from the training sample in a recursive manner \cite{charbuty2021classification}. It is an iterative process whereby data is partitioned into partitions and then further partitioned on each branch. The features used at different nodes are selected using statistical methods like Information Gain or Gini Index. If all features are already used and the remaining sample contains instances belonging to more than one class a leaf is created and its class will be decided using a majority vote.

\subsection*{5.5 Random forest}

The Random Forest (RF) method \cite{breiman2001random} was developed to overcome a shortcoming of Decision Trees, i.e., having the tendency to overfit the sample data. RF mitigates this problem by using a statistical technique called bootstrapping, which generates multiple models and combines their results to make a final decision. The main idea behind RF is that by aggregating the predictions of multiple classifiers, the impact of individual errors can be minimized.

In course of Bootstrapping several smaller samples are drawn from the training dataset randomly with replacement. Each sample is used to train a separate classifier. Thus when classifying a new observation
its final class prediction is made by aggregating the results given by the individual models. Usually, it is done by applying a majority voting solution.

\section*{6 Training the classifiers - Experimental results}
 
\begin{figure*}[!ht]
    \centering
    \includegraphics[width=1\textwidth]{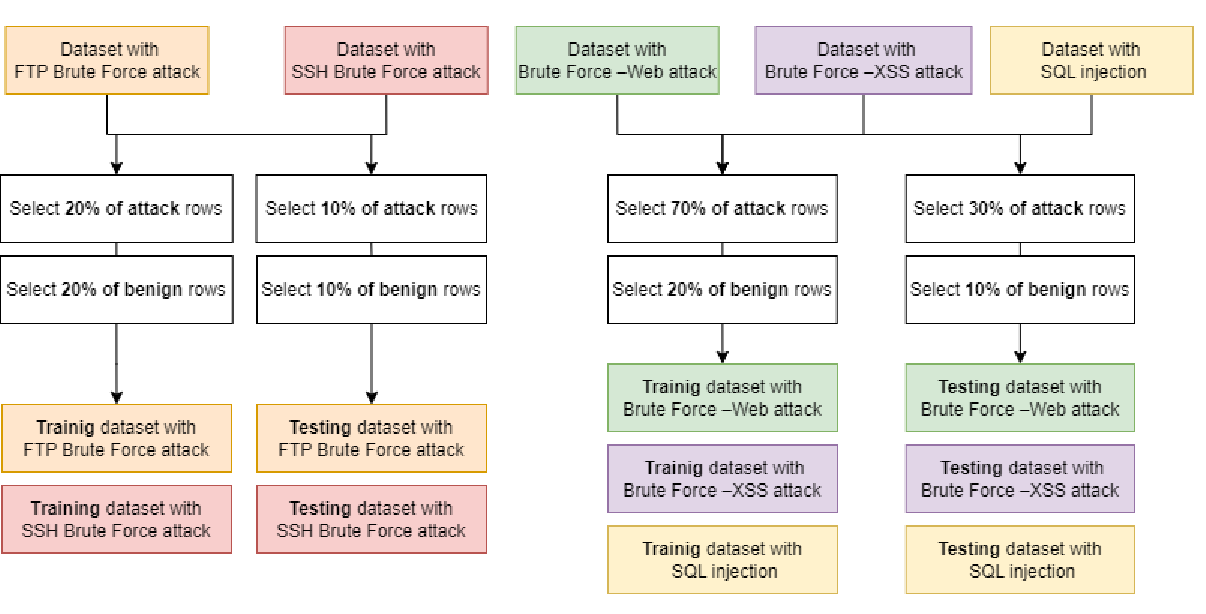}
    \caption{Creation of training and testing datasets}
    \label{fig:train_test_dataset_creation}
\end{figure*}

All five datasets contained a very large number of instances (see Table~\ref{table:Datasets_preproc}). Therefore when creating the training and test samples only a part of the original data was used. The steps of the training-test sample construction are presented in Fig.~\ref{fig:train_test_dataset_creation}.

Both in the case of the FTP and SSH Brute Force attacks training samples contained 20\% of the original instances. We applied stratified sampling to ensure that each class (attack and benign traffic) is represented in the sample. Thus the resulting collection of records contained 20\% of the attack rows and 20\% of the rows describing benign traffic. The sampling was carried out without replacement. The test samples were created in a similar way by choosing records from the remaining datasets so that the resulting collection of data points represented 10\% of the original ones. The resulting record numbers are shown in Tables \ref{table:training_datasets} and \ref{table:test_datasets}, respectively. 

In the case of the Brute Force Web, Brute Force XSS, and SQL Injection attack types the selection process was slightly different owing to the fact that the number of records describing malicious traffic was very small. Therefore in each case, the collection of attack rows was split into two parts, i.e.,  70\% was used for training and the remaining 30\% for test purposes. Next, the training samples were created by adding 20\% of the data points belonging to the benign traffic. Finally, the test samples were compiled by adding 10\% of the benign traffic record to the attack rows allocated for test purposes. The resulting record numbers are shown in Tables \ref{table:training_datasets} and \ref{table:test_datasets}, respectively. 

\begin{table}[!ht]
\renewcommand{\arraystretch}{2}
\caption{Training datasets}
\label{table:training_datasets}
\begin{tabular}{p{0.28\columnwidth}p{0.25\columnwidth}p{0.31\columnwidth}}
\hline
\textbf{File name} &\textbf{Number of rows} &\textbf{Number of columns}\\ 
\hline
dataset-ftp-tr.csv     & \hfil \num{171433}   & \hfil \num{69} \\
dataset-ssh-tr.csv     & \hfil \num{170280}   & \hfil \num{69} \\
dataset-web-tr.csv     & \hfil \num{417332}   & \hfil \num{69} \\
dataset-xss-tr.csv     & \hfil \num{417218}   & \hfil \num{69} \\
dataset-sql-tr.csv     & \hfil \num{417042}   & \hfil \num{69} \\
\hline
\end{tabular}
\end{table}

\renewcommand{\arraystretch}{2}
\begin{table}[!ht]
\caption{Test datasets}
\label{table:test_datasets}
\begin{tabular}{p{0.28\columnwidth}p{0.25\columnwidth}p{0.31\columnwidth}}
\hline
\textbf{File name} &\textbf{Number of rows} &\textbf{Number of columns}\\ 
\hline
dataset-ftp-ts.csv     & \hfil \num{85716}    & \hfil \num{69} \\
dataset-ssh-ts.csv     & \hfil \num{85140}    & \hfil \num{69} \\
dataset-web-ts.csv     & \hfil \num{208636}   & \hfil \num{69} \\
dataset-xss-ts.csv     & \hfil \num{208598}   & \hfil \num{69} \\
dataset-sql-ts.csv     & \hfil \num{208517}   & \hfil \num{69} \\
\hline
\end{tabular}
\end{table}

The training and testing of the five classifiers was carried out in Orange 3.34, which is an  open-source data visualization, machine learning, and data mining toolkit. It offers a visual programming front-end for interactive data visualization and exploratory, quick qualitative data analysis. Its components are called widgets and they range from simple data visualization, subset selection, and preprocessing to empirical evaluation of learning algorithms and predictive modeling. Visual programming is implemented through an interface in which workflows are created by linking predefined or user-designed widgets, while advanced users can use Orange as a Python library for data manipulation and widget alteration.
Orange uses common Python open-source libraries for scientific computing, such as numpy, scipy, and scikit-learn, while its graphical user interface operates within the cross-platform Qt framework. 

\begin{figure*}[!ht]
    \centering
    \includegraphics[width=1\textwidth]{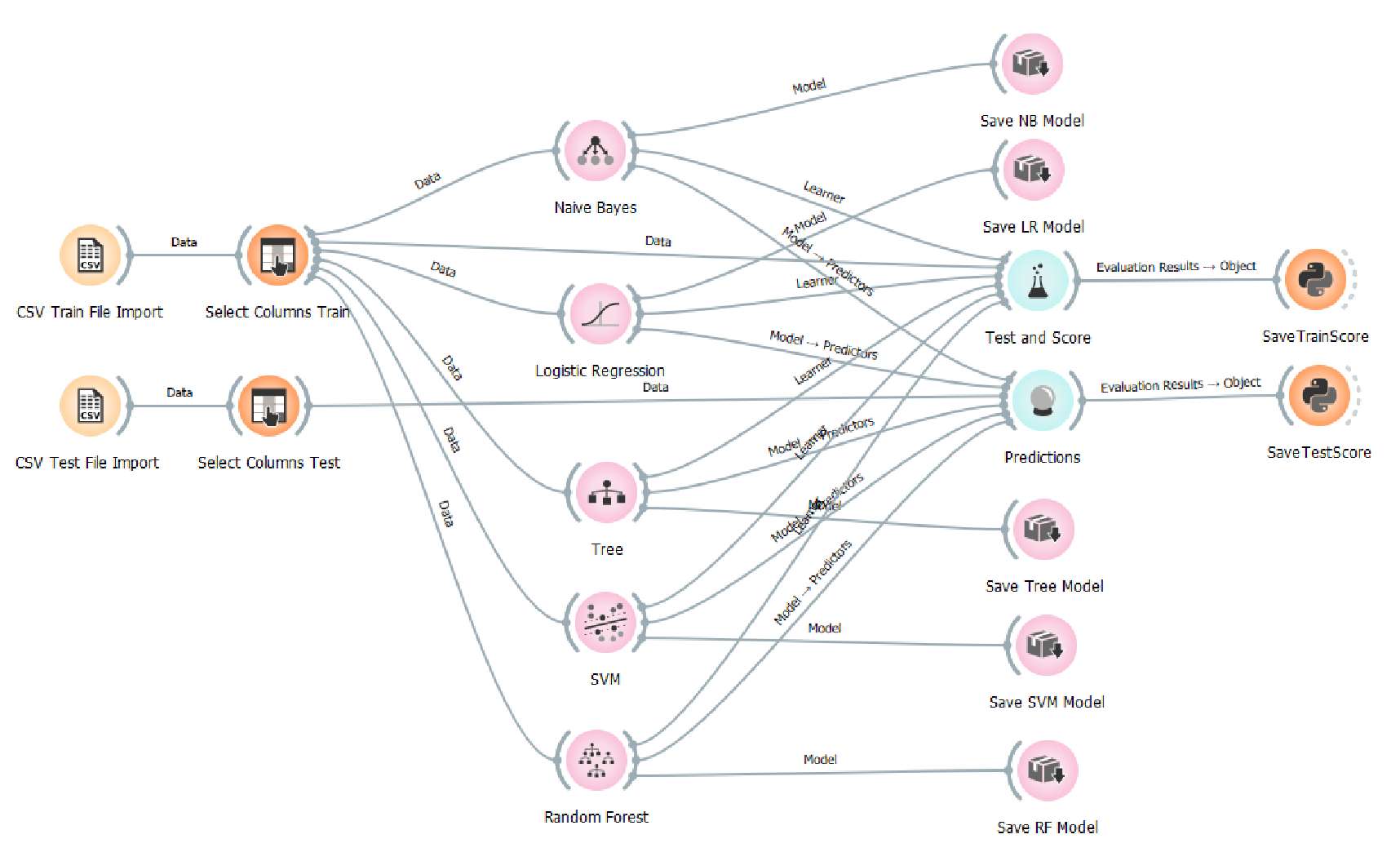}
    \caption{Classifier training and testing workflow}
    \label{fig:orange_workflow}
\end{figure*}

The classifier training and testing workflow used in course of the investigation is shown in Figure \ref{fig:orange_workflow}. It was carried out separately for each attack type and for each relevant feature collection. For example, in the case of the FTP Brute Force attack and the 0.40 ranking threshold value three features (44, 56, and 59) were supposed to play a significant role. Thus in total 21 workflow executions were necessary and 105 classifiers were trained.

All the classifiers were evaluated against the training and test samples using four measures, i.e., accuracy, precision, recall, and F1. The detailed results can be found in the Appendix in Tables A6 - A10. 

\begin{figure}[h!]
    \centering
    \includegraphics[width=0.5\columnwidth]{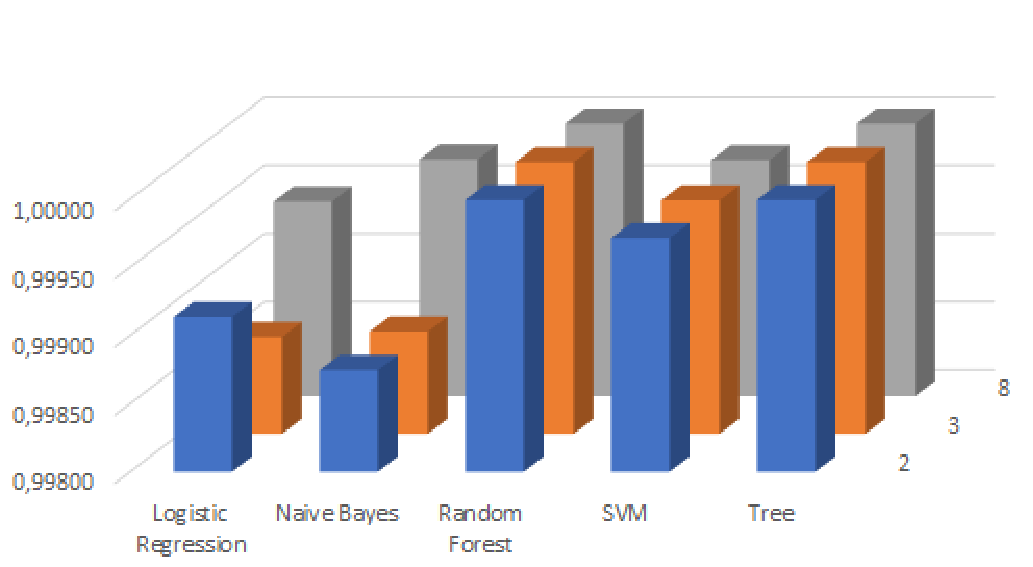}
    \caption{Accuracy values in case of the FTP attack and the training dataset}
    \label{fig:FTP_train_Accuracy}
\end{figure}

\begin{figure}[h!]
    \centering
    \includegraphics[width=0.5\columnwidth]{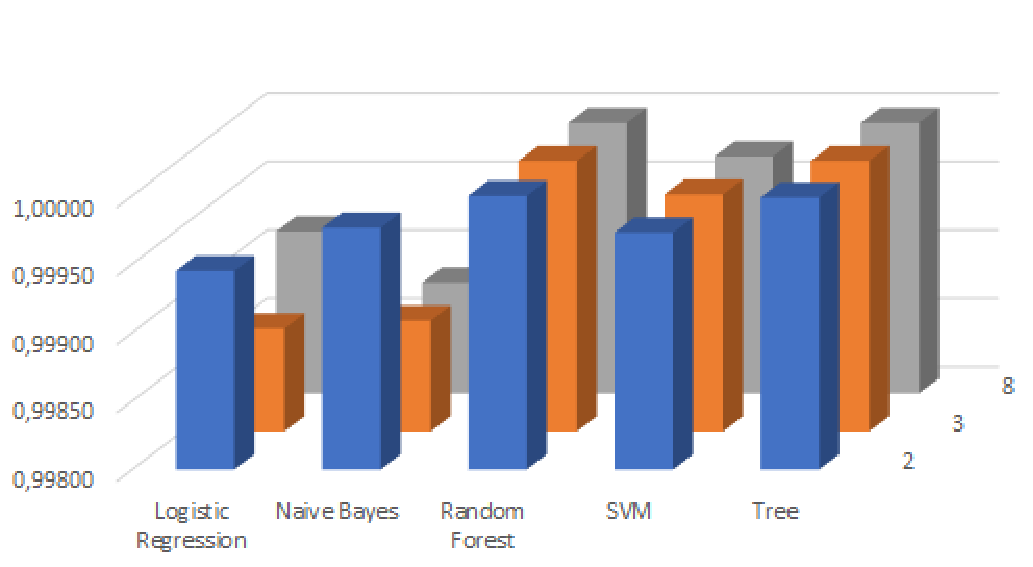}
    \caption{Accuracy values in case of the FTP attack and the test dataset}
    \label{fig:FTP_test_Accuracy}
\end{figure}

In the case of FTP attacks each of the classifiers performed quite well having high accuracy values and the highest possible recall rates for almost all of the feature subset-classifier type pairs. Except for the case of the logistic regression-based classifier, the accuracy against the training dataset usually improved with increasing the number of selected features (see Fig~\ref{fig:FTP_train_Accuracy}). However, when evaluating the classifiers with the test dataset a slight decay in accuracy performance could be measured in the case of Naive Bayes and Random Forest classifiers as well (see Fig~\ref{fig:FTP_test_Accuracy}).

\begin{figure}[h!]
    \centering
    \includegraphics[width=0.5\columnwidth]{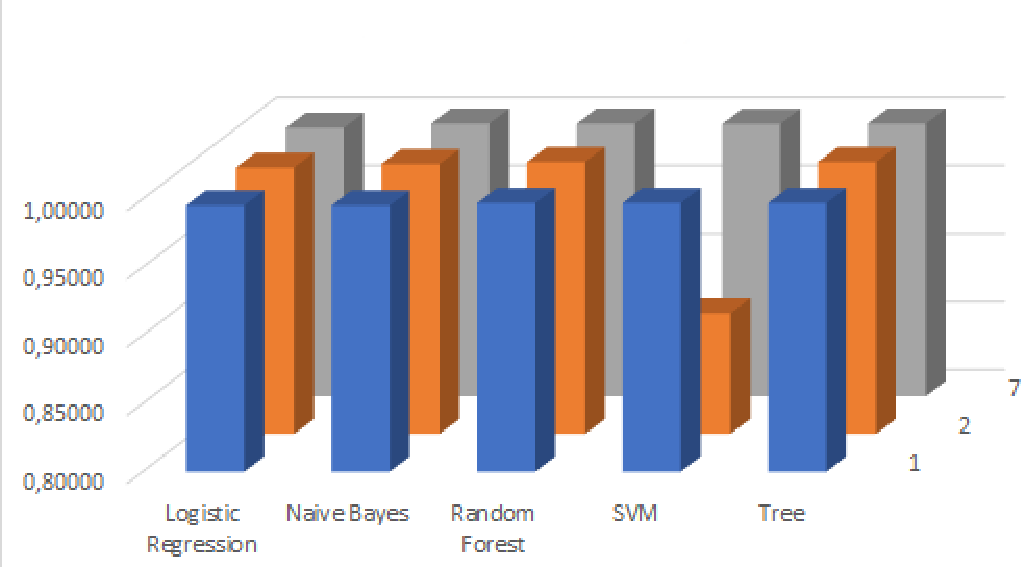}
    \caption{Accuracy values in case of the SSH attack and the training dataset}
    \label{fig:diagram_ssh_train}
\end{figure}

\begin{figure}[h!]
    \centering
    \includegraphics[width=0.5\columnwidth]{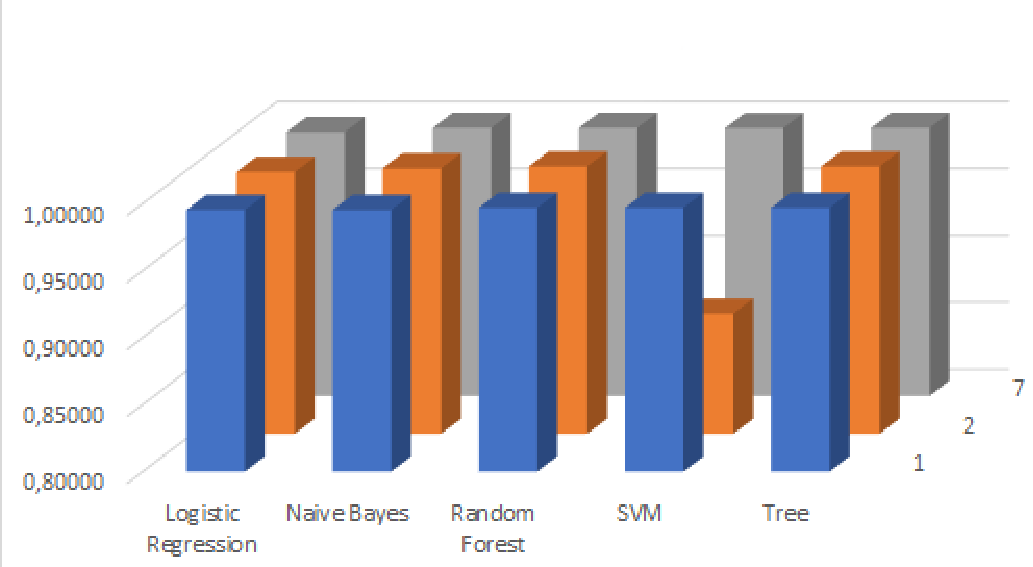}
    \caption{Accuracy values in case of the SSH attack and the test dataset}
    \label{fig:diagram_ssh_test}
\end{figure}

The classifiers also exhibited strong performance against SSH attacks, with high accuracy observed for all feature subset-classifier pairs. Increasing the number of selected features led to improved accuracy against the training dataset, except in the case of the SVM-based classifier (see Fig~\ref{fig:diagram_ssh_train}). When evaluating the classifiers against the test dataset revealed a very similar behavior (see Fig~\ref{fig:diagram_ssh_test}).

\begin{figure}[h!]
    \centering
    \includegraphics[width=0.5\columnwidth]{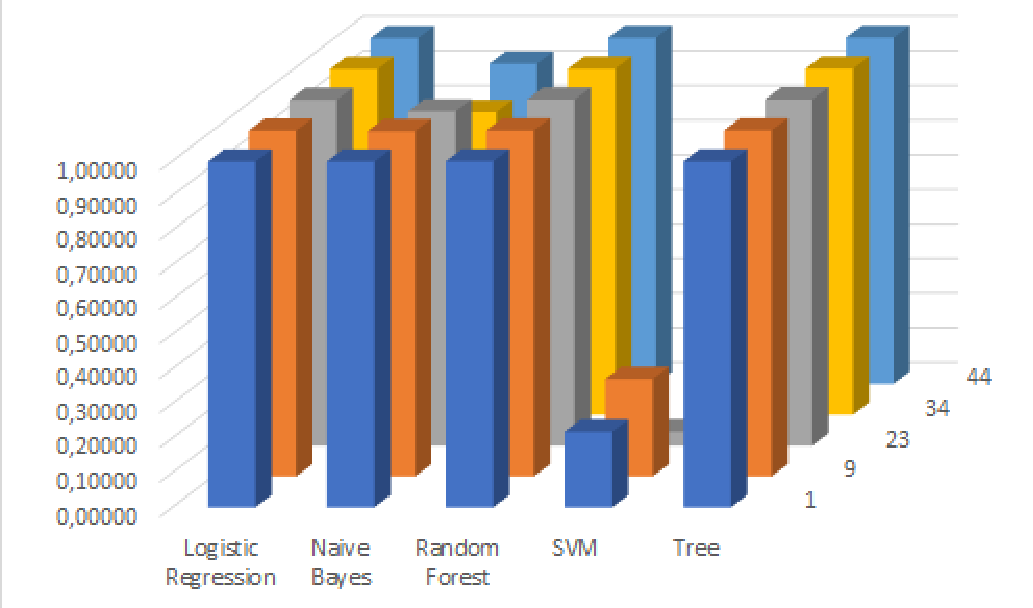}
    \caption{Accuracy values in case of the WEB attack and the training dataset}
    \label{fig:diagram_web_train}
\end{figure}

\begin{figure}[h!]
    \centering
    \includegraphics[width=0.5\columnwidth]{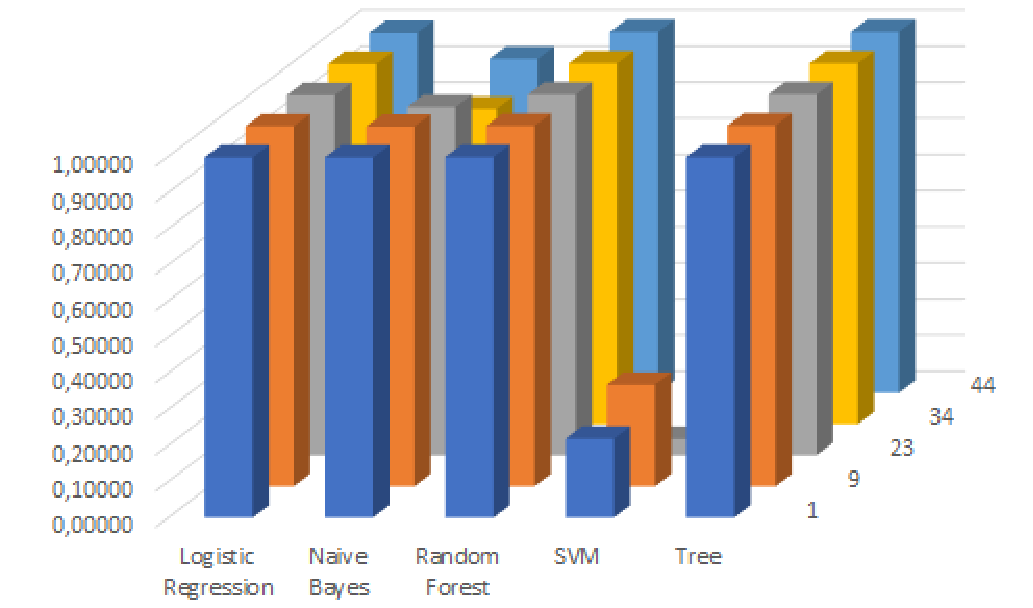}
    \caption{Accuracy values in case of the WEB attack and the test dataset}
    \label{fig:diagram_web_test}
\end{figure}

In the case of Web attacks, the SVM classifier provided a low accuracy rate compared to the others both in the case of the training (see Fig.~\ref{fig:diagram_web_train}) and test (see Fig.~\ref{fig:diagram_web_test}) datasets. However, Logistic regression, Random Forest, and Decision Tree based classifiers were able to successfully predict the nature of the traffic with a very high accuracy rate. Although the Naive Bayes model in the case of 23 and 34 selected features showed a declining performance its results were not too much fallen behind.  

\begin{figure}[h!]
    \centering
    \includegraphics[width=0.5\columnwidth]{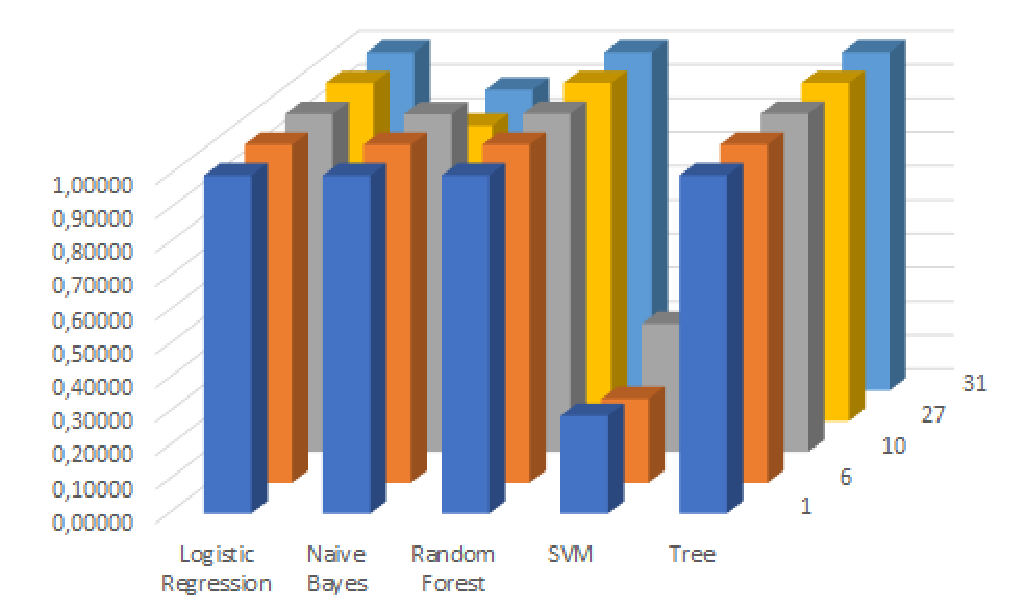}
    \caption{Accuracy values in case of the XSS attack and the training dataset}
    \label{fig:diagram_xss_train}
\end{figure}

\begin{figure}[h!]
    \centering
    \includegraphics[width=0.5\columnwidth]{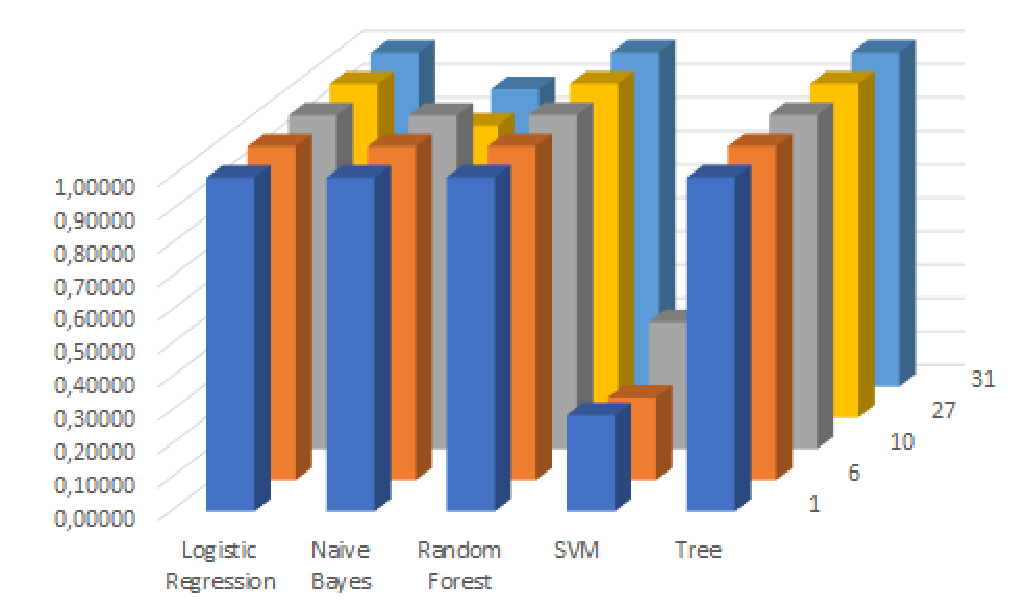}
    \caption{Accuracy values in case of the XSS attack and the test dataset}
    \label{fig:diagram_xss_test}
\end{figure}

Among the classifiers tested for XSS attacks, the SVM classifier had a low accuracy rate for both the training dataset (see Fig.\ref{fig:diagram_xss_train}) and test dataset (see Fig.\ref{fig:diagram_xss_test}) in comparison to others. The Logistic Regression, Random Forest, and Decision Tree based classifiers performed exceptionally well with a very high accuracy rate. While the Naive Bayes model showed declining performance in the case of 27 and 31 selected features, its results were still competitive.

\begin{figure}[h!]
    \centering
    \includegraphics[width=0.5\columnwidth]{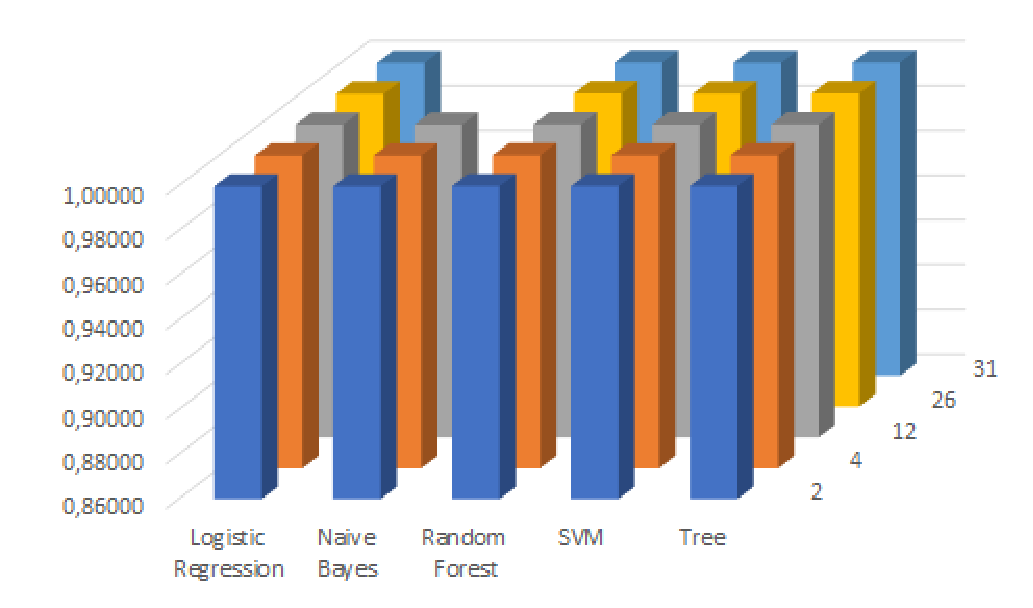}
    \caption{Accuracy values in case of the SQL attack and the training dataset}
    \label{fig:diagram_sql_train}
\end{figure}

\begin{figure}[h!]
    \centering
    \includegraphics[width=0.5\columnwidth]{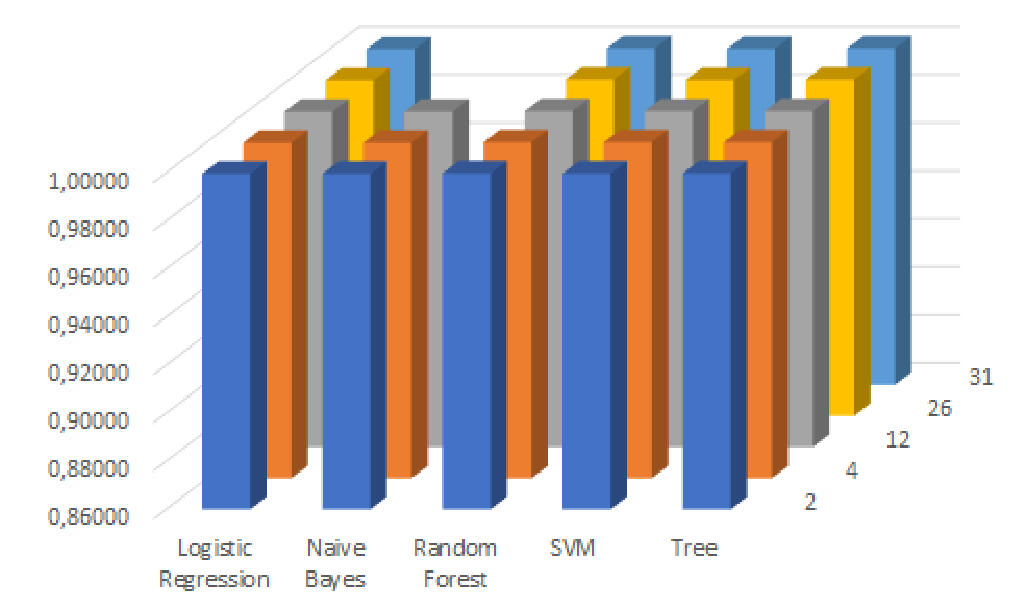}
    \caption{Accuracy values in case of the SQL attack and the test dataset}
    \label{fig:diagram_sql_test}
\end{figure}

For SQL Injection attacks, four of the five classifier models demonstrated high accuracy rates against both the training dataset (see Fig.~\ref{fig:diagram_sql_train}) and test dataset (see Fig.~\ref{fig:diagram_sql_test}). The Naive Bayes model was the only exception, showing slightly declining performance when using 26 and 31 selected features, but still having accuracy values over 0.9.

The best-performing classifiers along with ranking threshold values and selected feature number as well as the evaluation results are highlighted in Table~\ref{table:Best_performing_classifiers}

\begin{table*}[h!]
\renewcommand{\arraystretch}{2}
\caption{Best performing classifiers }
\label{table:Best_performing_classifiers}
\begin{tabular}{p{0.1\textwidth}p{0.1\textwidth}p{0.1\textwidth}p{0.13\textwidth}p{0.1\textwidth}p{0.1\textwidth}p{0.1\textwidth} p{0.07\textwidth}}

\hline
\hfil \textbf{Attack Type} &\hfil \textbf{Ranking Treshold} & \hfil \textbf{Features } &\hfil \textbf{Classifier}  & \hfil \textbf{Accuracy} & \hfil \textbf{Precision}   & \hfil \textbf{Recall}  &\hfil \textbf{F1}    \\ \hline
\hfil FTP & \hfil0.35 & \hfil 8  & Random Forest     & \hfil 1.00000  & \hfil 1.00000  & \hfil1.00000 & \hfil1.00000  \\
\hfil SSH & \hfil0.35 & \hfil 7  & Random Forest     & \hfil 0.99999  & \hfil 0.99997  & \hfil1.00000 & \hfil0.99999 \\
\hfil WEB & \hfil0.35 & \hfil 44 & Tree              & \hfil 0.99994  & \hfil 0.98997  & \hfil0.96890 & \hfil0.97932 \\ 
\hfil XSS & \hfil0.45 & \hfil 10 & Random Forest     & \hfil 0.99999  & \hfil 1.00000  & \hfil0.97391 & \hfil0.98678 \\
\hfil SQL &\hfil 0.40  & \hfil 26 & Tree              &\hfil  0.99999  & \hfil 1.00000  & \hfil0.95402 & \hfil0.97647\\\hline
\end{tabular}
\end{table*}

\section*{7 Conclusions}

In course of the investigation reported this paper, six feature evaluation techniques were performed on five datasets after completing the data cleaning and transformation steps. Each dataset comprised records that described two types of traffic cases: benign and attack, and featured 69 attributes. An average score was calculated after normalization and used to rank individual features. Next, six ranking thresholds were defined, which led to the selection of several relevant feature collections for each attack type. The number of included attributes varied widely, ranging from 1 (SSH) to 44 (Web).

Next, five classifier models were trained for each collection using the Orange software tool, and their performance was evaluated against the train and test datasets using four classification metrics. It was observed that, in some cases, accuracy improved slightly when increasing the number of features. However, excellent results were achieved in most cases, even with a low number of attributes. Table~\ref{table:Best_performing_classifiers}, which shows the best-performing classifiers for each attack type, clearly indicates that no general threshold can be set for the feature scores. The results suggest that tree-type classification algorithms represent the most appropriate solution for the investigated attack types when feature selection followed the presented workflow.

The methodology utilized in the current investigation can be applied to other scenarios involving high-dimensional data, such as clustering (e.g. \cite{blavzivc2019incremental}\cite{borlea2021unified}),  object identification \cite{hvizdovs2015object}, classification \cite{pokoradi_classification_2017}, indor localization \cite{Vincze}, or technology optimization \cite{babivc2019new}.

Further research will focus on the investigation of the suitability of different aggregation techniques (e.g. \cite{lilik2022fuzzy}\cite{toth2012effect}) that could replace the average score in feature relevance calculation. Furthermore the applicability of further computational intelligence methods.

\section*{Acknowledgements}
 On behalf of the project we are grateful for the possibility to use ELKH Cloud \cite{Heder_2022}; (\href{https://science-cloud.hu/}{https://science-cloud.hu/}), which helped us achieve the results published in this paper.

 This research was supported by 2020-1.1.2-PIACI-KFI-2020-00062 "Development of an industrial 4.0 modular industrial packaging machine with integrated data analysis and optimization based on artificial intelligence, error analysis". The Hungarian Government supports the Project and is co-financed by the
European Social Fund.







\nocite{*} 
\bibliographystyle{ios1}           
\bibliography{bibliography}        

%

\end{document}